\documentclass[conference]{IEEEtran}
\IEEEoverridecommandlockouts
\usepackage{cite}
\usepackage{amsmath,amssymb,amsfonts}
\usepackage{algorithmic}
\usepackage{graphicx}
\usepackage{textcomp}
\usepackage{booktabs}
\usepackage{subcaption}

\PassOptionsToPackage{hyphens}{url}\usepackage[breaklinks=true,bookmarks=false]{hyperref}
\usepackage[table,xcdraw,rgb]{xcolor}
\usepackage{tikz}
\usepackage{nicefrac, xfrac}
\usepackage{mathtools}
\usepackage{array,multirow,graphicx}
\usepackage{pifont}
\def\BibTeX{{\rm B\kern-.05em{\sc i\kern-.025em b}\kern-.08em
    T\kern-.1667em\lower.7ex\hbox{E}\kern-.125emX}}
\usepackage{fancyhdr}

\newcommand{\yes}{\ding{51}}%
\newcommand{\skeleton}{\includegraphics[height=2.5mm]{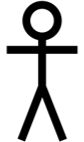}} 
\newcommand{\object}{\includegraphics[height=2.0mm]{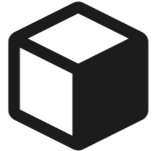}}
    
\begin{document}

\newcommand\scalemath[2]{\scalebox{#1}{\mbox{\ensuremath{\displaystyle #2}}}}

\renewcommand*{\figureautorefname}{Fig.}
\renewcommand*{\tableautorefname}{Tab.}
\renewcommand*{\equationautorefname}{Eq.}
\renewcommand*{\sectionautorefname}{Sec.}
\renewcommand*{\subsectionautorefname}{Sec.}
\title{%
Including Semantic Information via Word Embeddings
for Skeleton-based Action Recognition
}

\author{\IEEEauthorblockN{Dustin Aganian, Erik Franze, Markus Eisenbach, and Horst-Michael Gross}
\IEEEauthorblockA{\textit{Ilmenau University of Technology, Neuroinformatics and Cognitive Robotics Lab} \\
98684 Ilmenau, Germany \\
\small\tt dustin.aganian@tu-ilmenau.de, ORCID: 0009-0006-3925-6718
}}

\maketitle

\newcommand\todo[1]{\textcolor{red}{{#1}}}
\newcommand\redtext[1]{\textcolor{red}{{#1}}}

\newboolean{isarxiv}
\setboolean{isarxiv}{false}
\ifthenelse{\boolean{isarxiv}}{%
    \renewcommand{\headrulewidth}{0pt}
    \fancypagestyle{fancyfirstpage}{%
        \fancyhf{}%
        \fancyhead[C]{%
            \scriptsize%
               \textcolor{lightgray}{%
                  \small%
                   This work has been submitted to the IEEE for possible publication.\\
                   Copyright may be transferred without notice, after which this version may no longer be accessible.%
               }%
        }
        \fancyfoot[C]{%
            \footnotesize%
            \textcolor{gray}{\thepage}%
        }
    }
    \fancypagestyle{fancypage}{%
        \fancyhf{}%
        \fancyfoot[C]{%
            \footnotesize%
            \textcolor{gray}{\thepage}%
        }
    }
    \thispagestyle{fancyfirstpage}
    \pagestyle{fancypage}
}{%
    \thispagestyle{empty}%
    \pagestyle{empty}%
}%

\begin{abstract}
Effective human action recognition is widely used for cobots in Industry 4.0 to assist in assembly tasks.
However, conventional skeleton-based methods often lose keypoint semantics, limiting their effectiveness in complex interactions.
In this work, we introduce a novel approach to skeleton-based action recognition that enriches input representations by leveraging word embeddings to encode semantic information.
Our method replaces one-hot encodings with semantic volumes, enabling the model to capture meaningful relationships between joints and objects.
Through extensive experiments on multiple assembly datasets, we demonstrate that our approach significantly improves classification performance, and enhances generalization capabilities by simultaneously supporting different skeleton types and object classes.
Our findings highlight the potential of incorporating semantic information to enhance skeleton-based action recognition in dynamic and diverse environments.
\end{abstract}
\section{Introduction} \label{sec:Introduction}
In the context of Industry 4.0, the collaboration between humans and collaborative robots (cobots) is becoming increasingly popular for implementing manufacturing processes \cite{matheson2019human, sherwani2020collaborative, inkulu2021challenges}.
To effectively assist in assembly processes, cobots must first possess the capability to visually perceive the worker and recognize the current assembly state.
A critical component of achieving this objective is human action recognition \cite{eisenbach2021, terreran2023skeleton}.

To achieve effective generalization across diverse environments and individuals, skeleton-based action recognition emerges as a superior alternative to image-based approaches.
A recent study \cite{aganian2023object} demonstrated that incorporating object information into skeleton-based approaches is a viable strategy to further improve recognition capabilities.
However, to ensure seamless integration with neural networks, keypoints of the skeleton and the objects must be encoded appropriately.
A common approach involves representing the spatial position of each keypoint on a separate heatmap and subsequently stacking all heatmaps.
However, this approach results in the loss of semantic information associated with each keypoint. 

\begin{figure}[!t]
    \centering
    \includegraphics[width=1\columnwidth]{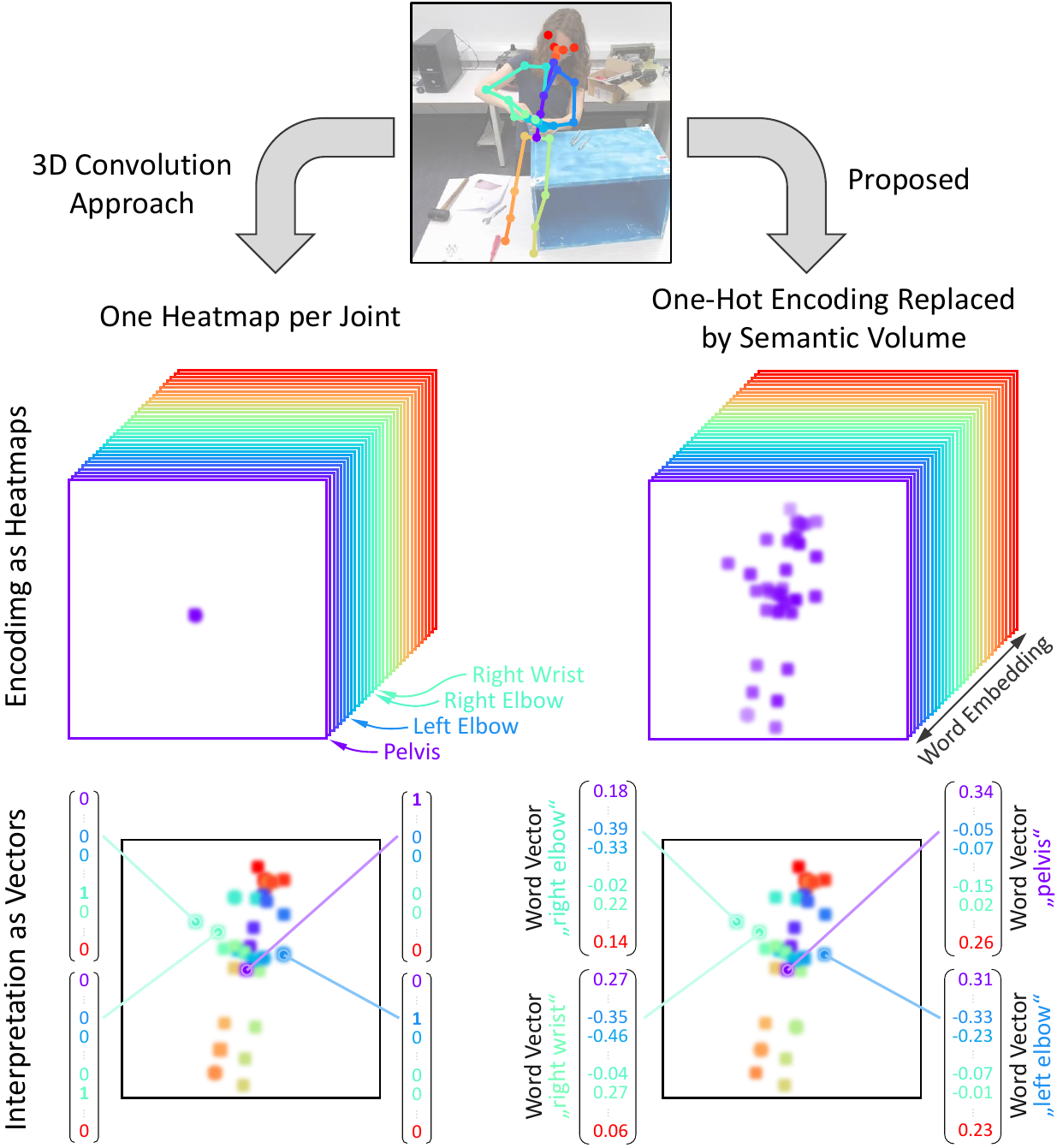}
    \vspace*{-5mm}%
    \caption{A semantic input encoding can be used to replace the conventional heatmap-based encoding of keypoints:
    At each time step, the skeleton must be encoded so that it can be processed by a 3D CNN.
    A common approach involves encoding the position of each joint of the skeleton in a separate heatmap, which can be interpreted as encoding each joint with a one-hot vector.
    Substituting the one-hot encoding of skeleton joints enables the application of word embeddings to encode the joints.
    This additional semantic information contributes to the improvement of skeleton-based action recognition.}
    \label{fig:BasicPrinciple}
    \vspace*{-6mm}%
\end{figure}

To encode more semantic information into the skeleton input, our approach expands upon the methods of \cite{PoseConv3D-cvpr2022, aganian2023object}, which use 3D convolutions to process sequences of 2D skeletons and optionally also object information.
The basic idea of using a semantic volume as keypoint encoding with semantic information is presented in \autoref{fig:BasicPrinciple}.
In the following, we briefly present the baseline based on \cite{PoseConv3D-cvpr2022, aganian2023object}, discuss its limitations, and then explain how our novel approach addresses and overcomes these issues.

\emph{Baseline}: 
The original input encoding of \cite{PoseConv3D-cvpr2022} is a 4D tensor for a skeleton sequence with the dimensions $J \times T \times H \times W$, where $J$ is the number of distinct joint classes of the skeleton, $T$ the count of frames in the sequence, and $H$ and $W$ refer to the height and width of the cropped frame.
The skeleton joints for a single time frame are encoded as heatmaps -- one heatmap for each joint class -- which are stacked along the joint/keypoint axis to describe a heatmap tensor ($J \times H \times W$).
An example can be seen on the left side of \autoref{fig:BasicPrinciple}.
In~\cite{aganian2023object}, this input encoding has been extended to not only use a skeleton detector, but also an object detector to detect relevant object data (e.g., tools, workpieces) in the frame to improve the detection and generalization capabilities of skeleton-based action classification.
The detected object data can then be used -- similar to skeleton data -- by interpreting the center of mass for each object as a further keypoint and append the additional heatmaps for every object class to the input for each frame.
This input is subsequently processed by a 3D CNN backbone.

Although these state-of-the-art methods achieve good results on test datasets, they exhibit several limitations that we aim to address with our approach.

First, encoding data as a heatmap volume results in significant loss of semantic information.
For example, information about the left elbow is encoded in the seventh heatmap, the right elbow in the 14th heatmap, and the right wrist in the 15th heatmap.
The neural network must learn these relationships from scratch, as the heatmaps are essentially one-hot encoded.
This is inefficient, considering the semantic relationships between joints are known prior to training.

Second, these models are typically tied to a specific skeleton type with a fixed number of joints.
For example, a model trained on Kinect Azure skeletons cannot be directly applied to a Mask R-CNN skeleton without significant degradation in performance.
This lack of flexibility hinders generalization across datasets and applications, which is particularly valuable during training and deployment on diverse data sources.

Third, integrating object data into the framework further exacerbates this issue.
Models are rigidly trained on specific object classes, requiring complete retraining with a new input size whenever classes are added or replaced.
This lack of adaptability is a significant barrier in dynamic environments.

Finally, the size of the input and associated memory requirements pose a scalability challenge.
As the number of skeleton joints or object classes increases, the size of the heatmap tensors for each frame grows substantially, leading to a dramatic rise in memory demands of the 3D CNN.

We aim to address these issues with our semantic volume, which is introduced in the following.

\emph{Semantic Volume Model Motivation}:
The core idea is to replace the one-hot encoding composed of individual heatmaps by introducing a semantic volume, where semantic information is encoded as word embeddings.
A visual example of this approach is shown on the right side of \autoref{fig:BasicPrinciple}.
This innovation aims to resolve all previously mentioned challenges.
We use the names of skeletal joints or objects as the basis for the word embeddings.
With an appropriately designed embedding, these feature vectors exhibit meaningful similarities.
For instance, joints that are semantically similar have a higher cosine similarity compared to object class names, as illustrated in \autoref{fig:similaritiesWordVectors16dAnd300d}.
Semantic relationships, such as similarities between vectors, are thus inherently encoded in the input.

Additionally, this approach allows different skeleton types to be used simultaneously as input.
The input size is no longer dependent on the number of skeletal joints, but is instead determined by the fixed length of the word embeddings.
Consequently, the input size remains unchanged regardless of whether object data are included or how many object classes are used.
Therefore, the memory requirements are no longer tied to the increasing number of heatmaps but are directly linked to the chosen length of the word embeddings.
This ensures a more scalable and efficient solution.

In \autoref{sec:sota}, we provide an overview of related work.
Then, in \autoref{sec:Approach}, we present a comprehensive exposition of the pipeline of our model and an in-depth discussion of the construction of the employed word embeddings.
Our experiments in \autoref{sec:Experiments} offer empirical evidence that the proposed input encoding results in a substantial enhancement in performance when compared to the baseline, which lacks semantic information in the input encoding.

Our code is publicly available at:
\href{https://github.com/TUI-NICR/semantic-embeddings-for-action-recognition}{https://github.com/TUI-NICR/semantic-embeddings-for-action-recognition}

\section{Related Work}
\label{sec:sota}
As outlined in \autoref{sec:Introduction}, our approach aims to encode additional semantic information into the input to address and improve various challenges in skeleton-based action recognition for sequences.
To achieve this, we leverage word embeddings of joint names and, optionally, object names in the sequence.
This allows us to encode relationships between joints and objects in the input with minimal additional computational overhead.

To the best of our knowledge, no existing method in the state of the art directly aligns with our approach.
The only related works we are aware of, which attempt to provide additional semantic information to skeleton-based action classifiers, include methods like \cite{aganian2023object, xu2022skeleton}.
These approaches combine skeleton data with object information, as previously mentioned.
Alternatively works like \cite{diller2024CVPR} focus on long-term future prediction and action forecasting.
Semantic information is gathered for forecasting by encoding and embedding past action outputs alongside skeletons and objects to better predict future sequences.
However, these methods do not explicitly encode semantic relationships between skeleton joints and objects; instead, the network must learn these relationships implicitly.

While these methods derive semantic information for downstream tasks, they do not directly enhance skeleton-based action recognition.

Consequently, we now present the state of the art for all the individual components that constitute our approach.

\subsection{Skeleton-based Action Recognition}
Skeleton-based action recognition for sequences typically involves processing temporal and spatial information from skeleton data.
Common approaches include 2D CNNs \cite{VACNN-TPAMI2019}, Transformers \cite{mazzia2022action-transformer}, GCNs \cite{Shi2019AGCN}, RNNs \cite{du2015rnn-action-recognition}, and 3D CNNs \cite{PoseConv3D-cvpr2022}.
For our approach, 3D CNNs and Transformers seem to be well-suited for handling spatiotemporal skeleton data enhanced with semantic word vectors.
We ultimately chose 3D CNNs due to the availability of a strong baseline \cite{PoseConv3D-cvpr2022, aganian2023object} for our target scenario, which allows us to highlight the improvements brought by our method.
Additionally, \cite{aganian2023object} incorporates object information as additional keypoints, a strategy we also adapt for our approach to include semantic information via word embeddings.

\subsection{Word Embeddings for Semantic Encoding}
To encode additional semantic information, we leverage word embeddings.
Our requirements for these embeddings are twofold:
\begin{itemize}
    \item They must encode rich semantic information through distances and similarities between words.
    \item They must have a fixed dimensionality to ensure compatibility with our approach.
\end{itemize}

Dynamic embeddings, such as BERT \cite{devlin2019bert} or ChatGPT \cite{achiam2023gpt}, are unsuitable for our task because they are context-dependent and do not provide word-based representations, which are essential for maintaining a consistent input structure for our approach.
Instead, we focus on static embeddings such as Word2Vec \cite{mikolov2013word2vec}, GloVe \cite{pennington2014glove}, and fastText \cite{bojanowski2017fasttext}.
Among these, we selected fastText due to its robustness and superior performance in capturing semantic relationships between words.

However, high-quality embeddings like fastText are typically high-dimensional (e.g., 300 dimensions), which poses significant memory challenges for our approach.
While some methods, such as GloVe, offer lower-dimensional variants (e.g., 50 dimensions), these still impose substantial memory overhead and suffer from reduced quality due to the large vocabulary it encodes (over 1 million words).

\subsection{Dimensionality Reduction for Word Embeddings}
To address this, we explore dimensionality reduction techniques that preserve semantic relationships while reducing the embedding size.
Many existing methods, such as t-SNE \cite{van2008tsne} or UMAP \cite{mcinnes2018umap} are unsuitable because they are mainly employed to reduce embeddings to 2 or 3 dimensions for visualization purposes, resulting in significant loss of semantic information.

Instead, we consider methods such as an extension of traditional PCA \cite{raunak2019pca-embeddings}, which removes unnecessary dimensions while preserving semantic relationships.
Additionally, some approaches \cite{chia2022reduce-vocab} reduce vocabulary size before simplifying or reducing the dimensionality of word vectors.
These methods often focus on selecting which words to remove from the vocabulary.

For our work, we combine these ideas:
We manually curate a small, task-relevant vocabulary using the lexical database WordNet \cite{miller1995wordnet} and then reduce the dimensionality of the word vectors while preserving the relationships within the selected vocabulary. 
In \autoref{sec:ApproachWordEmbedding}, we present a simple neural encoder for this purpose and compare it to the above mentioned PCA variant \cite{raunak2019pca-embeddings}.

\section{Including Semantic Information via Word Embedding} \label{sec:Approach}

\begin{figure*}[htb]
    \centering
    \includegraphics[width=1\textwidth]{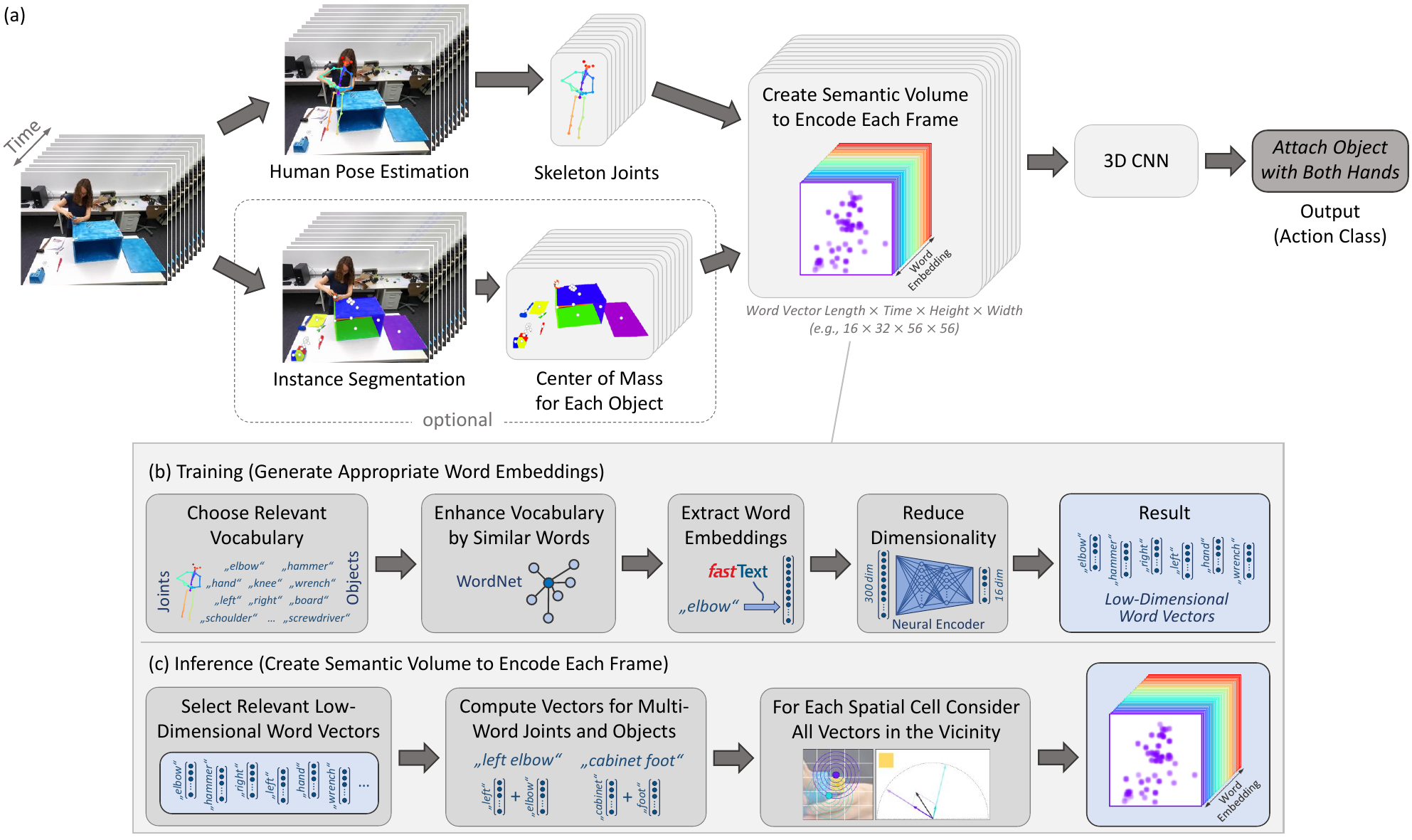}
    \vspace*{-6mm}%
    \caption{Processing pipeline of our approach for skeleton-based action recognition with semantic input encoding.
    (a) The skeletons and objects are extracted for each time frame.
    Then, they are encoded so that they can be processed by a 3D CNN that predicts the action class.
    Central to our approach is the encoding of the skeletal joints and the center of mass of the objects into a semantic volume for each time frame.
    (b) The process of encoding requires learning low-dimensional word vectors for all relevant joint and object class names.
    (c) These word vectors are then employed to encode the semantic information at each spatial position in the semantic volume.}
    \label{fig:pipeline}
    \vspace*{-5mm}%
\end{figure*}

\begin{figure*}[htb]
    \centering
    \includegraphics[width=1\textwidth]{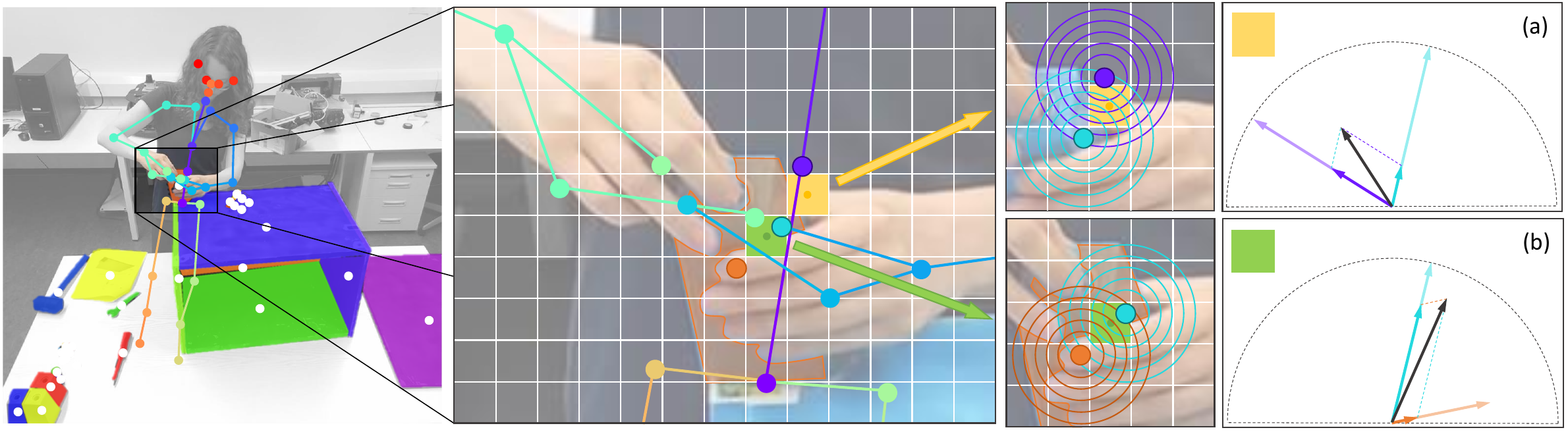}
    \vspace*{-6mm}%
    \caption{Combination of relevant semantic information for each spatial position in the semantic volume. The zoomed-in image shows a relevant part of the full image where the action of attaching an object with both hands is performed. The white grid resembles the actual resolution of the semantic volume. For each grid cell all relevant semantic information must be combined. Two situations are exemplarily shown: (a) The combination of two word vectors of skeleton joints ("spine navel", "left thumb") that are both close to the cell, and (b) the combination of word vectors of a skeleton joint ("left thumb") and those of an object ("cabinet foot") detected with a confidence score of 0.6 due to occlusions. The circles surrounding the keypoints indicate the multiples of the standard deviation ($1.0\sigma, 1.5\sigma, 2.0\sigma, 2.5\sigma, 3.0\sigma$) of Gaussians employed to weight the word vectors based on their distance to the cell. The sketches in (a) and (b) depict the actual angles between the 16-dimensional word vectors in a simplified 2D illustration.}
    \label{fig:overlay}
    \vspace*{-5mm}%
\end{figure*}

As introduced earlier, we build upon \cite{PoseConv3D-cvpr2022}, \cite{aganian2023object} and extend these methods by incorporating additional semantic information into the skeletal input using word embeddings.

We begin by presenting the baseline model without semantic information. Next, we describe how this baseline is extended to our semantic volume model. Finally, we explain the process of constructing the word embeddings required for our approach.

\subsection{Baseline}
Our baseline builds on \cite{PoseConv3D-cvpr2022, aganian2023object}, which are state-of-the-art methods for action classification using skeleton sequences, optionally enhanced with object information. The most distinctive aspect of these methods is their input encoding: skeleton data is represented as a heatmap volume stacked over time, enabling subsequent processing by a 3D CNN backbone. In the referenced works, a SlowOnly architecture \cite{slowonly-iccv2019}, derived from a direct inflation of ResNet-50, is employed as the 3D CNN. For consistency and easier comparison, we retain this architecture in our work.

As described in \autoref{sec:Introduction} and illustrated in \autoref{fig:BasicPrinciple}, the input is a $4D$ tensor representing a skeleton sequence to be classified, with dimensions $J \times T \times H \times W$.
Here, $T$ corresponds to a fixed temporal length, to which all skeleton sequences are mapped. $H$ and $W$ represent the height and width ($56 \times 56$) of the heatmap containing the skeleton, while $J$ denotes the number of joint classes in the skeleton.
In our experiments, we use two skeleton types, consisting of either $17$ or $32$ joints.

Following \cite{aganian2023object}, the input can also be augmented with object information.
This is achieved by applying an instance segmentation model, in addition to the skeleton detector, to each frame. Specifically, we predict object masks for relevant classes using a Mask R-CNN with a Swin Tiny Transformer backbone \cite{swin-transfomer-iccv2021}.
For all predicted object masks with a confidence threshold greater than $0.1$, the centers of mass of the detections are treated analogously to skeleton joints and encoded into corresponding heatmaps, weighted by their confidence scores.

Concretely, this means that for our input volume, the $J$ dimension is optionally extended by the number of relevant detected object classes.
For the datasets introduced in \autoref{sec:Datasets}, this results in an extension of $7$ or $12$ additional object classes.

The heatmap for the $j^{th}$ joint class at time $t$ is computed using Gaussian kernels centered at all estimated instances of the $j^{th}$ joint at time $t$, scaled by their corresponding confidence scores.
For a single estimated instance of the $j_i^{th}$ joint with position $(c_x^{(j_i)}, c_y^{(j_i)})$ and confidence score $s^{(j_i)}$, the resulting heatmap K is defined as follows:
    \vspace*{-1mm}%
\begin{equation}
    K_{j_i,x,y} = e^{-\frac{\left(x-c_x^{(j_i)}\right)^2 + \left(y-c_y^{(j_i)}\right)^2}{2 \cdot \sigma^2}} \cdot s^{(j_i)},
    \vspace*{-1mm}%
\end{equation}
where $\sigma$ denotes the variance of the Gaussian and is set to $0.6$ (as in \cite{PoseConv3D-cvpr2022}, \cite{aganian2023object}). %
During our experiments, we also tested other values for $\sigma$ ($0.2-0.65$) but gained no improvements.

    \vspace*{-1mm}%
\subsection{Semantic Volume Model}
    \vspace*{-1mm}%
The general workflow of our approach is illustrated in \autoref{fig:pipeline}a.
The key innovation lies in the creation of semantic volumes, where we replace the traditional "one-hot encoding" of heatmaps with a semantic volume, as depicted in \autoref{fig:BasicPrinciple}.
For semantic information, we use dimensionally reduced word embeddings of joint and object names, normalized to a feature vector length of approximately $1$.
Details are provided in the next subsection.

The resulting volume of stacked heatmaps can be interpreted as follows:
At the spatial locations corresponding to joints or objects, the word embeddings are "punctured" through the volume.
Consequently, the length of the $J$ dimension in our input tensor corresponds to the length of the word embeddings, regardless of the number of joint classes in the skeleton or the number of object classes modeled in the input.
Consequently, the volumes remain heatmaps, as we continue to scale the joints using Gaussian kernels weighted by their respective confidence scores.

To further illustrate the validity of this approach, \autoref{fig:overlay} demonstrates the computation of a single heatmap cell influenced by multiple joints.
For simplicity, we consider only two joints at a time in the visualizations in \autoref{fig:overlay}, although more joints have an influence on the discussed cells.
We consider two cases:

\begin{enumerate}
    \item Case 1 (\autoref{fig:overlay}a): The yellow cell is influenced by two joints ("spine navel" and "left thumb") located at approximately equal distances.
    The circles around the points represent multiples of the standard deviation from the joint locations.
    Both joints are estimated with a confidence of $1.0$, and their corresponding word vectors (normalized to a length of approximately $1.0$) are plotted on a unit circle for simplicity.
    The resulting vector (black arrow) lies directionally between the two joints and has a length roughly half that of the individual vectors.
    This reflects the desired representation for the cell, as it has a similar distance to both joints in image space.
    
    \item Case 2 (\autoref{fig:overlay}b): The green cell contains the joint "left thumb" and is influenced by a more distant "joint" "cabinet foot", which has a lower confidence score of $0.6$ due to occlusion by the hands.
    As shown in the bottom right, the orange vector (representing "cabinet food") is scaled by its confidence, reducing its length.
    Consequently, the resulting black vector is only slightly shifted toward the orange vector, as the influence of "cabinet foot" is diminished by both its distance and lower confidence.
    This allows the word embedding of the green cell to be interpreted as primarily representing "left thumb" with a minor contribution from "cabinet foot".
\end{enumerate}

We emphasize that the angles between the feature vectors presented in \autoref{fig:overlay} are derived using the actual angle between the shown word embeddings employed in our approach.
Therefore, the shown cases represent realistic relationships between the relevant word vectors shown in a simplified 2D sketch.

The process of creating these embeddings for all relevant words is described in the following subsection.

\subsection{Word Embedding}\label{sec:ApproachWordEmbedding}
To achieve the semantic relationships described above, our method relies on high-quality word embeddings that encode meaningful semantic information through the distances or similarities between relevant words. Additionally, these embeddings must have a low dimensionality to avoid memory issues during the processing of semantic volumes.

However, state-of-the-art semantic word embeddings, such as fastText (300 dimensions), typically have high dimensionality because they are trained on large vocabularies (e.g., 2 million words) and require this dimensionality to model complex relationships.
To obtain compact yet effective word embeddings for our approach, we combine several techniques from the literature.
Specifically, we select a smaller, task-relevant vocabulary and reduce the dimensionality of pre-trained embeddings for this vocabulary.
This allows us to eliminate unnecessary relationships with words outside the selected vocabulary while preserving the semantic relationships between the words within it.

\subsubsection{Training Process}
The workflow for obtaining compact word embeddings is illustrated in \autoref{fig:pipeline}b.
The process begins with the selection of an initial vocabulary, which includes the names of joints from all available skeleton detection methods and, optionally, the names of object classes from our relevant datasets.
This vocabulary is then expanded using WordNet~\cite{miller1995wordnet}, a lexical database, to include several hundred or thousand additional relevant words.
The extended vocabulary allows for better generalization in case of slight differences in relevant objects and skeletal joints between data during training and deployment.

Experimentally, we found that a vocabulary of 1,000 words is sufficient when working exclusively with skeleton data, while a vocabulary of 100 words suffices when incorporating object data.
Further details and explanations on these experiments are provided in \autoref{sec:AblationStudies}.

For the selected vocabulary, we extract high-dimensional word embeddings using the pre-trained fastText model \cite{grave2018fasttext-vec}.
These embeddings are then reduced to low dimensionality using a neural encoder.
Experimental results indicate that dimensions of 16 and 20 yield the best performance, as discussed later in \autoref{sec:AblationStudies}.

As an alternative to our neural encoder, we also explored a PCA variant \cite{raunak2019pca-embeddings} for dimensionality reduction.
However, this approach produced inferior results compared to our neural encoding method.

\subsubsection{Neural Encoding for Dimensionality Reduction} 
The goal of our dimensionality reduction is to preserve the semantic relationships between word vectors in the selected vocabulary.
To evaluate this, we use the cosine similarity as metric.
Specifically, we compute a pairwise cosine similarity matrix for the 300-dimensional fastText embeddings and compare it to the corresponding matrix for the embeddings with reduced dimensionality.
The difference between these matrices serves as our loss function and can be described as follows for every pair of word vectors $(w_a, w_b)^{(i)}$ in our vocublary of size $N$:
    \vspace*{-2mm}%
\begin{equation} \label{eq:CosineLoss}
\mathcal{L} = 
\frac{1}{\binom{N}{2}} \sum\limits_{i=1}^{\binom{N}{2}} 
\scalemath{0.95}{
\left( \text{cos}\left((w_a, w_b)^{(i)}\right) - \text{cos}\left((\phi (w_a), \phi (w_b))^{(i)}\right) \right)^2
}
\text{,}
\end{equation}
\vspace*{-6mm}%
\begin{equation*}
\text{where} \quad \text{cos}(w_a, w_b) = \frac{w_a \cdot w_b}{\|w_a\| \cdot \|w_b\|}
    \vspace*{-2mm}%
\end{equation*}
and $\phi$ is the neural encoder.

We train a fully connected neural network, similar to the encoder part of an autoencoder, to minimize our loss.
The best-performing architecture consists of two hidden layers with 200 and 150 neurons, respectively, and ReLU activation.
Additionally, we incorporate Ring Loss~\cite{zheng2018ring-loss} to enforce the neural encoder to learn a uniform length of approximately 1.0 for the word vectors.
Further details on these hyperparameter experiments are provided in \autoref{sec:Experiments}.

While our loss function from \autoref{eq:CosineLoss} provides a useful measure of embedding quality, the ultimate evaluation criterion is the performance of the action classification task using these embeddings.
This is particularly important because comparing loss values across different vocabularies is unfair, as smaller vocabularies inherently simplify the problem.

\begin{figure*}[!t]
    \centering
    \begin{subfigure}{.5\textwidth}
        \includegraphics[width=\textwidth]{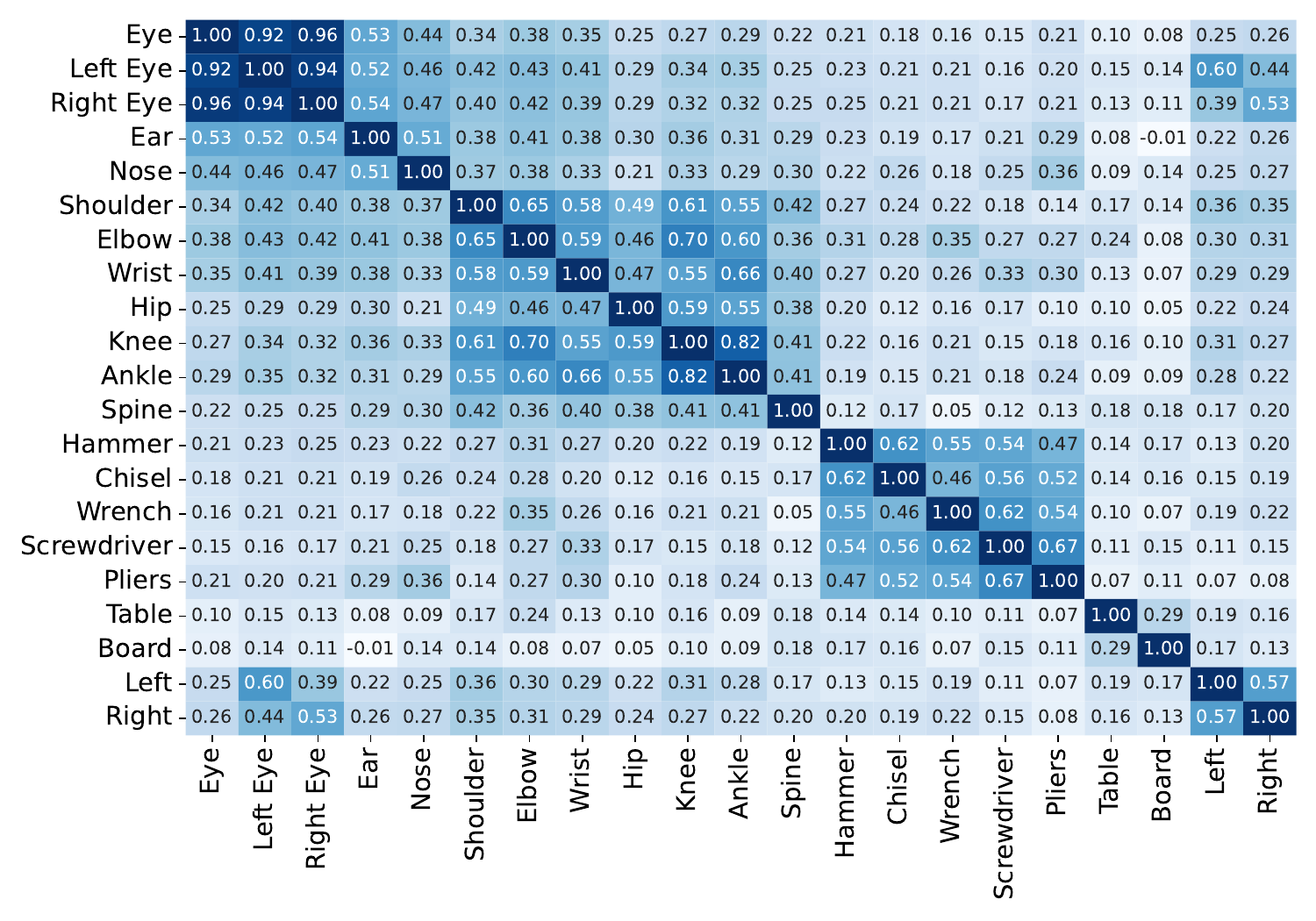}\\[-7mm]
        \caption{300-dimensional fastText word vector similarities}
    \end{subfigure}%
    \begin{subfigure}{.5\textwidth}
        \includegraphics[width=\textwidth]{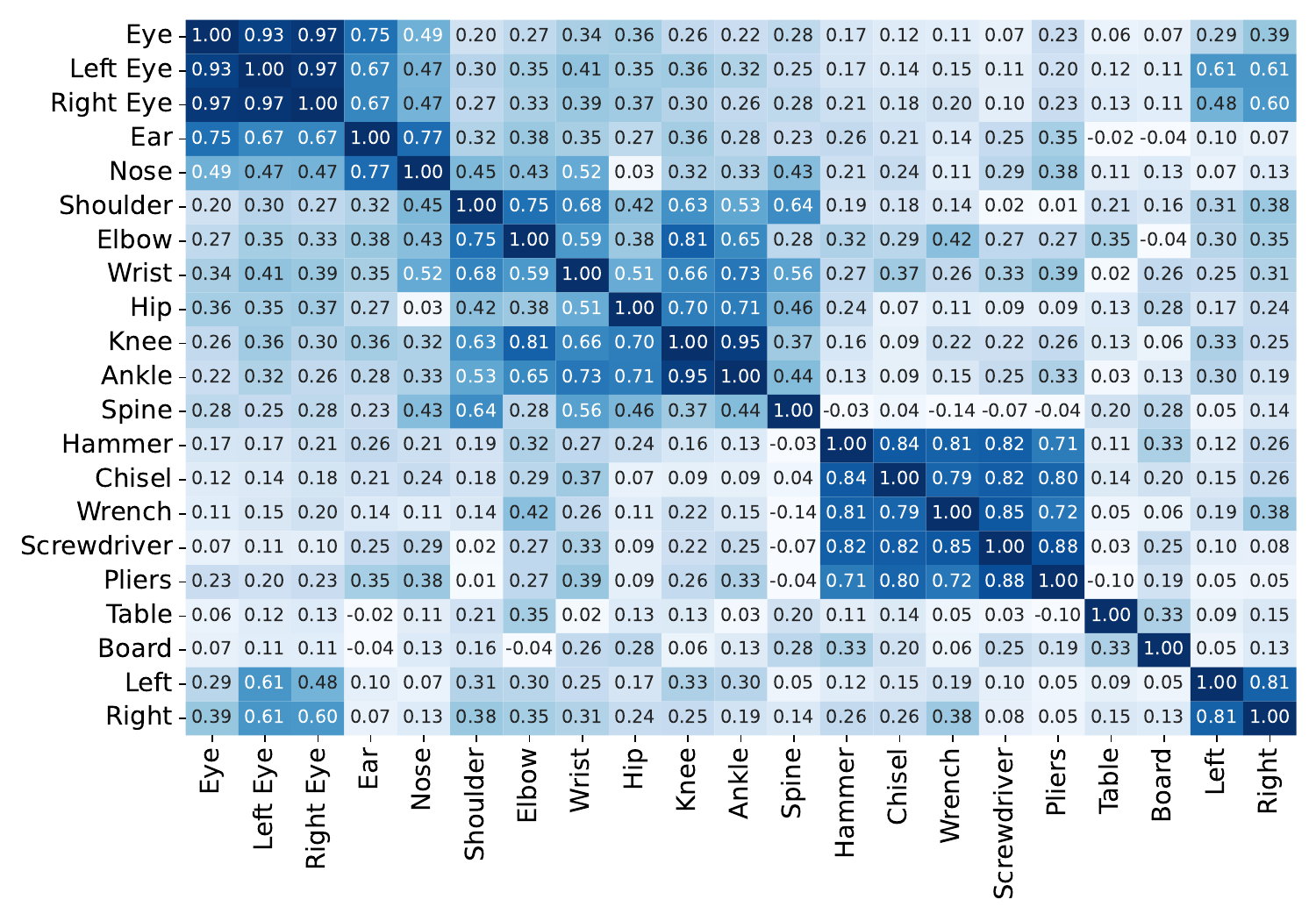}\\[-7mm]
        \caption{Encoded 16-dimensional word vector similarities}
    \end{subfigure}%
    \vspace*{-1mm}%
    \caption{Cosine similarities of word vectors for selected application-relevant vocabulary for (a) the original fastText vectors and (b) the low-dimensional word vectors learned by our neural encoder. The high degree of resemblance between the two similarity matrices indicates that semantic relationships are effectively preserved by the neural encoder.}
    \label{fig:similaritiesWordVectors16dAnd300d}
    \vspace*{-5mm}%
\end{figure*}

Finally, \autoref{fig:similaritiesWordVectors16dAnd300d} provides a visual comparison of the cosine similarity matrices for (a) the 300-dimensional fastText embeddings and (b) the 16-dimensional embeddings produced by our neural encoder.
The matrices are visually similar, and the computed loss according to \autoref{eq:CosineLoss} after training is approximately 0.01, indicating that the semantic relationships are well preserved.

\subsubsection{Inference for Semantic Volume Creation}
Finally, we explain how the dimensionally reduced word embeddings are used to create semantic volumes.
The overall process is illustrated in \autoref{fig:pipeline}c.

Step 1 --- Select Relevant Low-Dimensional Word Vectors:
The first step is self-explanatory: the pre-trained, dimensionally reduced word embeddings are selected for use.
    
Step 2 --- Compute Vectors for Multi-Word Joints and Objects:
Text embedders typically do not provide pre-trained word vectors for compound terms such as "left elbow" or "cabinet foot."
To address this, we combine the word vectors of the individual components.
While various approaches exist for combining word vectors, our experiments showed no significant differences in performance.
Therefore, we adopt a typical method: the dimensionally reduced word vectors are summed and then divided by the number of components in the compound term.
This approach approximates the creation of a word vector for a compound term using a space as the connecting character.

Step 3 --- Create Semantic Heatmap Volumes:
As preliminarily discussed in the case study for \autoref{fig:overlay}, the final step involves aggregating the word vectors while accounting for their spatial distribution using Gaussian kernels.
Specifically, the word vectors are weighted by their distance-based Gaussian contributions to compute the resulting feature vectors (black arrows in \autoref{fig:overlay}).
This process aligns with the heatmap volume creation described in \cite{PoseConv3D-cvpr2022}, \cite{aganian2023object}, but using semantic vectors instead of one-hot encoded vectors (see \autoref{fig:BasicPrinciple}).

However, this is not the only possible approach.
Given that the heatmaps are now "denser" due to the inclusion of semantic information, alternative aggregation methods may yield better results.
In our experiments, we explore two additional approaches:

\begin{itemize}
    \item Normalized Summation:
    The summed word vectors are divided by the number of Gaussian distributions influencing the cell.
    This prevents excessively high values in the heatmap when many joints or objects influence a single cell.
    \item Weighted Normalization:
    Instead of dividing by the number of Gaussian distributions, the division accounts for the distance and height of each Gaussian distribution.
    This ensures that the scaling of word vectors is explicitly considered, preventing overly small values in cases where many "distant" word vectors influence a cell.
\end{itemize}

These alternatives to simple summation are evaluated in detail in \autoref{sec:AblationStudies}.

\section{Experiments}\label{sec:Experiments}
In the following experiments, we first describe the employed datasets and the setup, then demonstrate the improvements achieved by our approach compared to the baseline.
Subsequently, we present ablation studies to analyze the impact of semantic information in the word vectors and the effects of various design parameters in our approach.
Finally, we showcase the generalization capabilities of our method by conducting transfer training across different skeleton types and object classes — a task that is not feasible with the original baseline approach.

\subsection{Datasets}\label{sec:Datasets}
In order to illustrate the efficacy of the proposed approach, we utilize two distinct datasets, \mbox{IKEA ASM}~\cite{IKEA-wacv2021} and \mbox{ATTACH}~\cite{attach}, in our experiments.
These datasets comprise sequences depicting furniture assembly and have been applied in previous work on skeleton-based action recognition \cite{aganian2023object, aganian2023fusing}.
The actions were captured from multiple perspectives, with skeletal data available for all viewpoints.
However, significant discrepancies exist between the two datasets.

The videos of IKEA ASM are recorded with Kinect V2 cameras and contain ca. 25,000 labeled action instances, distributed over 33 atomic classes.
The 48 participants assembled different furniture types in various environments, which makes the dataset challenging.
IKEA ASM provides 2D skeleton predictions from OpenPose \cite{openpose-cvpr2017} and Keypoint R-CNN.
We use the latter in our experiments due to its better performance.
The skeletons predicted by this method contain 17 joints.
Furthermore, IKEA ASM includes seven object classes, but data is not available for all perspectives.

The assembly processes of ATTACH are captured by Kinect Azure cameras, with 42 participants assembling cabinets on a table.
The resulting dataset contains ca. 95,000 annotated actions distributed across 51 classes.
In contrast to the IKEA ASM dataset, the actions are annotated separately for each hand, which can result in overlapping annotations, creating a significant challenge.
By employing the Azure Kinect body tracking SDK, 3D skeletons consisting of 32 joints were estimated for ATTACH.
In \cite{aganian2023fusing}, 2D skeletons were generated from the 3D data, which are utilized in this study.
Furthermore, on ATTACH, a distinction is made between 12 object classes, for which masks are provided \cite{eisenbach2024detection-fsod}, though not for all frames.

The original datasets lack the object data necessary for the proposed approach.
Consequently, a Mask R-CNN with a Swin Tiny Transformer backbone \cite{swin-transfomer-iccv2021} is employed to generate the required object masks, as previously mentioned in \autoref{sec:Approach}.
The object detector predicts masks for all relevant classes and is applied to each frame of both datasets.
As previously outlined, this methodology was initially employed for IKEA ASM in \cite{aganian2023object} and we have since adopted it for the ATTACH dataset.
Within the context of our approach, a threshold of 0.1 is applied to filter out predictions with a low confidence score.

\subsection{Setup} \label{sec:Setup}
For the implementation of our approach based on \cite{PoseConv3D-cvpr2022, aganian2023object}, we used MMAction2 \cite{2020mmaction2}.
In general, we oriented on the original hyperparameters and repeated the training of each configuration three to five times to ensure consistent results.
Regardless of the dataset, we used the SGD optimizer, a cosine annealing learning rate schedule and a maximum learning rate of $2.5 \times 10^{-2}$ in our trainings.
The number of training epochs is adjusted for each dataset. For IKEA ASM, the training consists of 240 epochs, whereas for ATTACH, it is limited to 60 epochs due to the larger amount of training data in ATTACH.
We validated after each epoch and performed testing according to the datasets protocols and the test protocol of \cite{PoseConv3D-cvpr2022} using the checkpoint for the best validation result.

For evaluation, we applied mean class accuracy (mAcc) as well as top-1 accuracy (top1) as these are often used in action recognition papers \cite{PoseConv3D-cvpr2022, aganian2023object, aganian2023fusing}.
The advantage of mAcc is that it accounts for class imbalances in the dataset, whereas for top1 all action clips contribute equally.
During hyperparameter optimization we consistently focused on maximizing mAcc rather than top1.
This choice was motivated by the significant class imbalance present in the two datasets used.
As a result, top1 values tend to fluctuate considerably, leading to outliers and reduced reproducibility of results.
By prioritizing mAcc, we ensure more robust and reliable performance metrics.

An extensive hyperparameter search was conducted for training the neural encoder.
This is easily feasible since the training can be performed without significant time constraints.
Various configurations were tested, including different dimensions, word vector lengths, vocabulary sizes, network topology, learning rates, activation functions, and learning rate schedulers.
The relevant best values we identified are detailed in \autoref{sec:ApproachWordEmbedding}, while the experimental results for dimensionality, word vector length, and vocabulary size are discussed further in \autoref{sec:AblationStudies}.

\subsection{Proof of Concept}

We begin the presentation of our results by comparing the performance of our baseline with that of our proposed approach.
This allows us to demonstrate that the inclusion of word vectors in the input not only appears theoretically sound, as previously discussed, but also yields measurable improvements in practice.

\begin{figure}[!h]
    \vspace*{-5mm}%
    \centering
    \includegraphics[width=1\columnwidth]
    {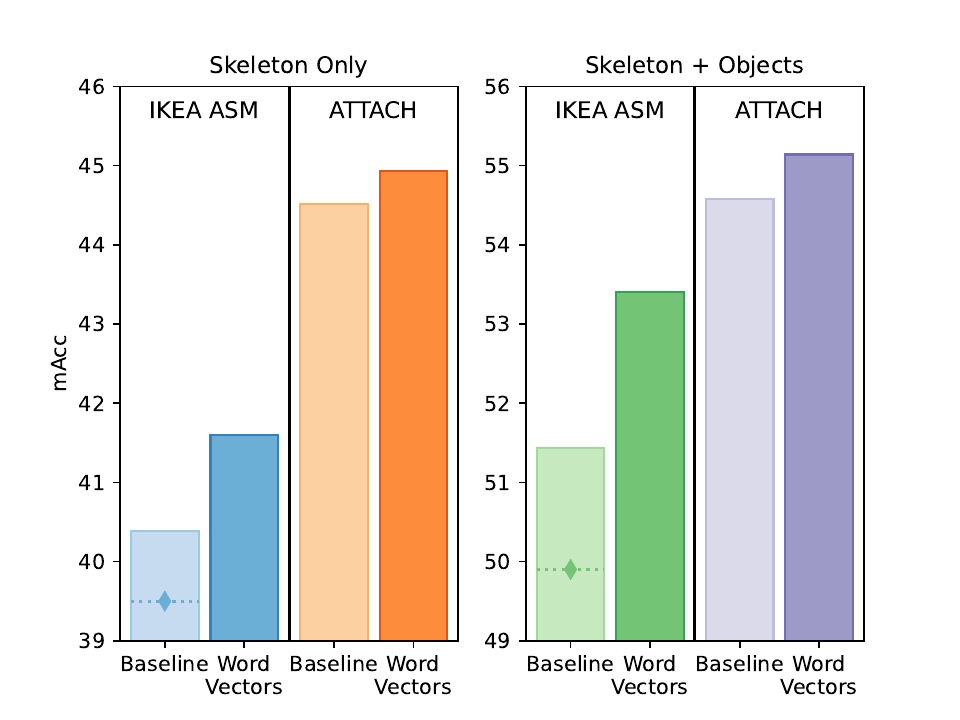}
    \vspace*{-6mm}%
    \caption{Best results in terms of mean accuracy (mAcc) on the IKEA ASM and the ATTACH dataset. The diamond in the IKEA ASM baseline bars shows the best results from literature~\cite{aganian2023object} with a method similar to our baseline, while the bars show the results with appropriate hyperparameter tuning.}
    \label{fig:resultsBarPlotAll}
    \vspace*{-2mm}%
\end{figure}

In \autoref{fig:resultsBarPlotAll}, we present the performance gains of our approach over the baseline.
Four experimental scenarios are compared: skeleton data only, and skeleton data combined with object detection results, evaluated on two assembly-based datasets, IKEA ASM and ATTACH (see \autoref{sec:Datasets}).

\subsubsection{Results on IKEA ASM}
\begin{table}[!b]
\centering
\caption{Results IKEA ASM dataset}
    \vspace*{-2mm}%
\label{tab:AllResultsIKEA}
\begin{tabular}{lccccc}
\toprule
Model                                    & Encoding     & \skeleton & \object & mAcc          & top1          \\ \midrule
\cite{aganian2023object} & One-Hot      & \yes      &                        & 39.5          & \textbf{75.0} \\
Ours                                      & One-Hot      & \yes      &                        & 40.4          & 72.7          \\
Ours                                      & Word Vectors & \yes      &                        & \textbf{41.6} & 74.7          \\ \midrule
\cite{aganian2023object} & One-Hot      & \yes      & \yes    & 49.9          & 79.7          \\
Ours                                      & One-Hot      & \yes      & \yes    & 51.4          & 79.2          \\
Ours                                      & Word Vectors & \yes      & \yes    & \textbf{53.4} & \textbf{82.1} \\ \bottomrule
\end{tabular}%
\end{table}

Our baseline corresponds to the same model used in \cite{aganian2023object} for both the skeleton-only and skeleton-object experiments, including appropriate hyperparameter tuning.
When comparing our baseline to \cite{aganian2023object} (see \autoref{tab:AllResultsIKEA}), we observe that our baseline performs better:
For skeleton-only experiments, we achieve a 0.9~percentage point improvement in mAcc, and for skeleton-object experiments, we achieve a 1.5~percentage point improvement in mAcc.
This improvement is likely due to our hyperparameter tuning focused on mAcc as stated in \autoref{sec:Setup}.

When comparing our proposed approach to our baseline, we observe additional gains of 1.2 and 2 percentage points in mAcc, further underscoring the advantages of our method.

\subsubsection{Results on ATTACH}

For ATTACH, we first establish a baseline.
Since the approach from \cite{PoseConv3D-cvpr2022, aganian2023object} has not been applied to ATTACH before, we present comparable results using 2D skeletons on a 2D ResNet-50 from \cite{aganian2023fusing} in \autoref{tab:AllResultsATTACH}.
This comparison demonstrates that the approach of \cite{PoseConv3D-cvpr2022} with a 3D ResNet-50-like network outperforms the approach of \cite{VACNN-TPAMI2019} with a 2D ResNet-50 on 2D skeletons.
Additionally, we reproduce the findings from \cite{aganian2023object} on a different dataset, showing that incorporating object data provides a significant performance boost.
Similar to the results on IKEA ASM, our approach outperforms the baseline on ATTACH.

\begin{table}[!h]
\centering
\caption{Results ATTACH dataset}
    \vspace*{-2mm}%
\label{tab:AllResultsATTACH}
\begin{tabular}{lccccc}
\toprule
Model                                    & Encoding     & \skeleton & \object & mAcc          & top1          \\ \midrule
ResNet-50 \cite{aganian2023fusing} & Skeleton Image \cite{VACNN-TPAMI2019}      & \yes      &                        & 43.7          & 52.3 \\
Ours                                      & One-Hot      & \yes      &                        & 44.5          & 54.7          \\
Ours                                      & Word Vectors & \yes      &                        & \textbf{44.9} & \textbf{55.1}          \\ \midrule
Ours                                      & One-Hot      & \yes      & \yes    & 54.6          & 61.3          \\
Ours                                      & Word Vectors & \yes      & \yes    & \textbf{55.1} & \textbf{61.5} \\ \bottomrule
\end{tabular}%
\end{table}

\subsubsection{Discussion}
These improvements on both datasets were achieved under identical hyperparameters and training conditions, respectively, solely through modifications to the input data.
This means that our approach incurs no additional computational overhead during inference.
In fact, the smaller input volumes (due to the typically smaller size of word vectors compared to one-hot encodings) reduce memory requirements.

This demonstrates that our approach not only offers the previously discussed advantages in terms of generalization properties but also achieves superior results on the respective datasets, both for skeleton-only and skeleton-object configurations.

After analyzing the final results of our approach in comparison to the state of the art and to our baseline, we present various ablation studies in the next section to examine different aspects of our approach in more detail.

\subsection{Ablation Studies}\label{sec:AblationStudies}
In the following, we will present the impact of various hyperparameters on the recognition performance and explain the rationale behind the design decisions in our approach, as outlined in \autoref{sec:Approach}.
It should be noted that, for all of our experiments, we ensured that all experiments within a ablation study were conducted under similar hyperparameter setups rather than focus on always using our final optimized hyperparameter set.
This approach was chosen to maintain comparability between the results and arose from the fact that various ablation studies were conducted at different stages during the development of our approach.
However, a drawback of this method is that the final test scores often do not reach the best performance reported in \autoref{fig:resultsBarPlotAll}.
Consequently, in the following analyses, we sometimes only present the relative differences in performance within each ablation study.
For simplicity and space constraints, we present results only on the IKEA ASM dataset.
However, we consistently observed the same trends and comparable results in our ablation studies on the ATTACH dataset as well.

\subsubsection{Impact of Semantic Information}
Notably, we begin this section with our investigation on how the experimental results compare when using random vectors instead of word vectors.
This analysis aims to demonstrate that the improvements are indeed due to the semantic information encoded in the word vectors, rather than merely a better initial utilization of the input space.

\begin{table}[htb]
\centering
\caption{mAcc for experiments with different vectors on IKEA ASM}
    \vspace*{-2mm}%
\label{tab:VectorStudy}
\begin{tabular}{llccccc}
\toprule
 \skeleton & \object & {\color[HTML]{9B9B9B} Baseline} & Random & Permutate & Switch & Approach        \\ \midrule
 \yes      &                        & {\color[HTML]{9B9B9B} 40.4}     & 39.9   & 38.8        & 40.3   & \textbf{41.6} \\
 \yes      & \yes    & {\color[HTML]{9B9B9B} 51.4}     & 50.7   & 51.1        & 52.3   & \textbf{53.4} \\ \bottomrule
\end{tabular}%
\end{table}

In \autoref{tab:VectorStudy}, we present the results of an ablation study comparing our approach with alternative vector types.
Alongside our baseline, which uses a one-hot encoding, we evaluate three other types of vectors that can be embedded into the heatmap volumes:

\begin{itemize}
    \item Random:
    The vectors for joint names are randomly initialized at the start of training, with a fixed length of 1.0 and the same dimensionality as the word vectors in our approach (16 dimensions). 
    This tests whether simply filling the input space more effectively is sufficient to improve performance.
    
    \item Permutate:
    The mapping between word vectors (from the neural encoder) and joint names is permuted before training.
    This tests whether the specific semantic relationships between joint names are necessary or if introducing some semantic structure into the input space is sufficient.
    
    \item Switch:
    The word vectors associated with human joints and object centroids are swapped.
    Unlike the \textit{permutate} approach, this method preserves the semantic relationships within each group (i.e., relationships between body parts and between objects, as shown in \autoref{fig:similaritiesWordVectors16dAnd300d}).
\end{itemize}

The results in \autoref{tab:VectorStudy} reveal that both the \textit{random} and \textit{permutate} versions perform worse than the baseline.
This demonstrates that neither better utilization of the input space nor the introduction of arbitrary semantic structure is sufficient to improve performance.

The \textit{switch} version yields interesting results:
It performs comparably to the baseline in the skeleton-only setting and even outperforms the baseline in the skeleton-object setting.
However, it still falls short of our proposed approach.
We hypothesize that the improvement in the skeleton-object setting arises because the network can better distinguish between body parts and objects due to the preserved intra-group similarities, providing an advantage over the one-hot encoding used in the baseline.

\subsubsection{Word Vectors}
Below, we discuss various experiments and design decisions regarding our word vector creation, which we introduced in \autoref{sec:ApproachWordEmbedding}.

\emph{Methods for Word Vector Dimensionality Reduction}:
We applied both our neural encoder approach and the PCA variant~\cite{raunak2019pca-embeddings} for reducing the dimensionality of word vectors.
Most of our experiments were initially conducted using PCA-reduced word vectors, but in this work, we only presented results obtained with our neural encoder.
The primary reason for this decision is that PCA consistently yielded inferior results, with a mAcc difference of approximately 0.5 to 2 percentage points, depending on the dataset and input data used.
This highlights the advantage of our simple yet effective approach, which trains a small fully connected network for optimized dimensionality reduction.
Another reason why the neural encoder outperformed PCA is its learned normalization, which seemed to play a relevant role in preserving useful information and creating better word vectors.

\emph{Optimal Length of Word Vectors}:
We evaluated different word vector lengths after dimensionality reduction.
In addition to the learned length of 1.0 through Ring Loss~\cite{zheng2018ring-loss} in the neural encoder, we tested word vectors where their length was either left unnormalized or explicitly normalized to 1.0 after training.
Our findings indicate that learned normalization via Ring Loss consistently performed better than fixed normalization, which in turn was slightly better than using unnormalized vectors.
The impact on mAcc was typically around 0.5 to 1 percentage point, highlighting the importance of proper length scaling.
Since word vectors are often summed during the creation of heatmaps, we also investigated shorter vector lengths.
However, no improvements were observed, and in some cases, performance even declined.

\emph{Optimal Dimensionality of Word Vectors}:
We further investigated the impact of different target dimensionalities for word vectors.
Specifically, we tested dimensions of 8, 12, 16, 20, and 24.
Our results show that dimensions of 16 and 20 yielded the best performance, while 12 and 24 also achieved nearly comparable results.
However, reducing the dimensionality to 8 resulted in significantly lower performance, with up to a 2.5 percentage point drop in mAcc.
This suggests that an excessively aggressive reduction leads to substantial loss of semantic information.
Based on these findings, we conclude that word vector dimensions should not be reduced below 12, while increasing them beyond 20 offers no substantial gains to justify the additional memory requirements.

\emph{Optimal Size of the Target Vocabulary}:
We tested vocabulary sizes ranging from fewer than 100 words to 10,000 words.
The optimal size depends on the dimensionality reduction method used.
For PCA, larger vocabularies (e.g., 5,000 words) yielded the best results.
In contrast, our neural encoder achieved optimal performance with smaller vocabularies of 100 or 1,000 words, provided the vocabulary was sufficiently diverse.

For skeleton-only experiments, we used WordNet to generate words similar to joint names, while for skeleton-object experiments, we included words related to both joints and objects.
In skeleton-object experiments, a vocabulary of 100 words was sufficient to achieve strong results, whereas skeleton-only experiments performed best with 1,000 words.
This demonstrates that, as long as the target vocabulary is sufficiently diverse, a smaller vocabulary is better suited for training the neural encoder.
We hypothesize that the neural encoder benefits from the stronger constraints imposed by the relationships within a diverse but compact vocabulary, leading to the formation of higher-quality word vectors.

Our target vocabulary of 100 words for assembly scenarios already includes all relevant terms, such as common tools, objects, and body parts, ensuring that our approach maintains strong generalization properties even with a compact vocabulary.

\subsubsection{Heatmap Volume Creation}
Finally, one more aspects of our approach requires further discussion.
We must address how intersecting word vectors should be combined during the creation of the heatmap, as mentioned at the end of \autoref{sec:ApproachWordEmbedding}.

\emph{Optimal Word Vector Aggregation}:
We evaluate three variants for combining intersecting word vectors:
Addition, normalized addition, and weighted normalization.
Preliminary results indicate that weighted normalization consistently underperforms in all experiments (1--3 percentage points mAcc worse).
We hypothesize that this is due to small vector magnitudes resulting from the weighted combination and the high influence of the weights on the angle of the resulting vector.

When comparing addition and normalized addition, the results differ between skeleton-only and skeleton-object experiments.
For skeleton-only experiments, addition and normalized addition yield nearly identical results.
However, for skeleton-object experiments, normalized addition performs significantly worse (up to 3 percentage points mAcc).
We attribute this to the increased number of keypoints in skeleton-object scenarios, due to detected objects, leading to more overlaps -- especially with low-confidence object detections.
Based on these findings, we decided in favor of the addition variant for our final approach.

\subsection{Cross-Dataset Transfer Learning}
\begin{table}[!b]
    \vspace*{-2mm}%
\centering
\caption{Results Transfer Learning}
    \vspace*{-2mm}%
\label{tab:TransferLearning}
\begin{tabular}{ccccccc}
\toprule
\multicolumn{3}{c}{}                                                          & \multicolumn{2}{c}{Pre Training}                           & \multicolumn{2}{c}{Results}                                                                 \\
\multicolumn{1}{l}{Dataset} & \multicolumn{1}{l}{\skeleton} & \multicolumn{1}{l}{\object}  & \multicolumn{1}{l}{IKEA ASM} & \multicolumn{1}{l}{ATTACH}  & \multicolumn{1}{l}{mAcc}              & \multicolumn{1}{l}{{\color[HTML]{9B9B9B} top1}}     \\ \midrule
                            & \yes                          &                             &                              &                             & \textbf{41.6}                         & {\color[HTML]{9B9B9B} 74.7}                         \\
                            & \yes                          &                             &                              & \yes                         & 40.9                                  & {\color[HTML]{9B9B9B} 74.6}                         \\
                            & \yes                          &                             & \yes                          & \yes                         & 39.6                                  & {\color[HTML]{9B9B9B} 74.1}                         \\
                            & \cellcolor[HTML]{EFEFEF}\yes  & \cellcolor[HTML]{EFEFEF}\yes & \cellcolor[HTML]{EFEFEF}     & \cellcolor[HTML]{EFEFEF}    & \cellcolor[HTML]{EFEFEF}53.4          & \cellcolor[HTML]{EFEFEF}{\color[HTML]{9B9B9B} 82.1} \\
                            & \cellcolor[HTML]{EFEFEF}\yes  & \cellcolor[HTML]{EFEFEF}\yes & \cellcolor[HTML]{EFEFEF}     & \cellcolor[HTML]{EFEFEF}\yes & \cellcolor[HTML]{EFEFEF}\textbf{54.5} & \cellcolor[HTML]{EFEFEF}{\color[HTML]{9B9B9B} 81.7} \\ 
\multirow{-6}{*}{\rotatebox[origin=c]{90}{IKEA ASM}}  & \cellcolor[HTML]{EFEFEF}\yes  & \cellcolor[HTML]{EFEFEF}\yes & \cellcolor[HTML]{EFEFEF}\yes  & \cellcolor[HTML]{EFEFEF}\yes & \cellcolor[HTML]{EFEFEF}51.3          & \cellcolor[HTML]{EFEFEF}{\color[HTML]{9B9B9B} 81.3} \\ \midrule
                            & \yes                          &                             &                              &                             & 44.9                                  & {\color[HTML]{9B9B9B} 55.1}                         \\
                            & \yes                          &                             & \yes                          &                             & \textbf{45.3}                         & {\color[HTML]{9B9B9B} 55.3}                         \\
                            & \yes                          &                             & \yes                          & \yes                         & 45.0                                  & {\color[HTML]{9B9B9B} 55.2}                         \\
                            & \cellcolor[HTML]{EFEFEF}\yes  & \cellcolor[HTML]{EFEFEF}\yes & \cellcolor[HTML]{EFEFEF}     & \cellcolor[HTML]{EFEFEF}    & \cellcolor[HTML]{EFEFEF}55.1          & \cellcolor[HTML]{EFEFEF}{\color[HTML]{9B9B9B} 61.5} \\
                            & \cellcolor[HTML]{EFEFEF}\yes  & \cellcolor[HTML]{EFEFEF}\yes & \cellcolor[HTML]{EFEFEF}\yes  & \cellcolor[HTML]{EFEFEF}    & \cellcolor[HTML]{EFEFEF}55.1          & \cellcolor[HTML]{EFEFEF}{\color[HTML]{9B9B9B} 61.9} \\
\multirow{-6}{*}{\rotatebox[origin=c]{90}{ATTACH}}    & \cellcolor[HTML]{EFEFEF}\yes  & \cellcolor[HTML]{EFEFEF}\yes & \cellcolor[HTML]{EFEFEF}\yes  & \cellcolor[HTML]{EFEFEF}\yes & \cellcolor[HTML]{EFEFEF}\textbf{55.8} & \cellcolor[HTML]{EFEFEF}{\color[HTML]{9B9B9B} 62.0} \\ \bottomrule
\end{tabular}
\end{table}

With our approach, we explore the newly gained capability of performing transfer learning across datasets with different skeleton types and object classes —  a feature that was previously impractical due to the effort and time required to align datasets and models, and as a result, is uncommon in related work on skeleton-based action recognition.
Since this is not the primary focus of this work, we tested only a simple, naive transfer learning approach.
We compared the performance of our approach trained on IKEA ASM with scenarios where the model was pre-trained on ATTACH or trained simultaneously on both ATTACH and IKEA ASM.
We also conducted the reverse experiment, with the goal of testing on ATTACH.
A summary of these experiments is provided in \autoref{tab:TransferLearning}.

In summary, transfer learning proved to be beneficial and provided improvements in three out of four cases.
Notably, pre-training on ATTACH before training on IKEA ASM for skeleton-object experiments yielded an improvement of more than 1 percentage point in mAcc.
This substantial gain could be attributed to the larger number of object classes in the ATTACH dataset, which enhances the model's generalization capabilities.
With more advanced transfer learning techniques and additional datasets, there is likely substantial room for further improvement, which can be explored in future work.

\section{Conclusion}
In the contemporary era of Industry 4.0, human-robot interaction demands sophisticated semantic understanding capabilities, including skeleton-based action recognition.
In this work, we introduced a novel approach for skeleton-based action recognition that enriches input representations by leveraging word vectors as semantic encoding.
Through extensive experiments, we demonstrated that this method significantly improves classification performance across multiple assembly datasets, surpassing both the baseline and state-of-the-art methods.
Our approach not only enhances the model's ability to generalize across diverse environments and individuals but also maintains computational efficiency by reducing the input volume compared to the typical one-hot encoding.

We conducted multiple ablation studies to analyze the key design choices behind our method.
By comparing word vectors to randomly assigned vectors, we confirmed that the performance gains stem from the semantic information embedded in word vectors rather than merely a better input space utilization.
Furthermore, we explored different techniques for word vector dimensionality reduction, and additional investigations into vector length, dictionary size, and heatmap volume creation provided further insights into the design choices of our approach.

Furthermore, our exploration of transfer learning demonstrated the potential of our method to adapt across datasets with different skeleton structures and object classes.
This highlights an exciting direction for future research, where further refinements and applications in new domains could unlock even greater benefits.

\bibliographystyle{IEEEtran}
\bibliography{literature_short.bib}

\end{document}